\documentclass{hotnets2020}
\usepackage{hyperref}
\hypersetup{pdfstartview=FitH,pdfpagelayout=SinglePage}

\usepackage[utf8]{inputenc} % allow utf-8 input
\usepackage{url}            % simple URL typesetting
\usepackage{booktabs}       % professional-quality tables
\usepackage{amsfonts,amsmath}       % blackboard math symbols
\usepackage{nicefrac}       % compact symbols for 1/2, etc.
\usepackage{microtype}      % microtypography
\usepackage{color}
\usepackage{graphicx}
\usepackage{xcolor}         % colors
\usepackage{colortbl}       % even more colors
\usepackage{fixltx2e}
\usepackage{subcaption}
\usepackage{cleveref}

\newcommand{\ProbName}{online safety assurance}
\newcommand{\ProbNameShort}{OSAP}

\title{Online Safety Assurance for Deep Reinforcement Learning}

\newcommand{\huji}{Hebrew University of Jerusalem}
\newcommand{\technion}{Technion}
\newcommand{\authemail}[1]{#1}

\begin{document}

\conferenceinfo{HotNets 2020} {}
\CopyrightYear{2020}
\crdata{X}
\date{}

\author{Noga H. Rotman \\ 
\huji \\ 
\authemail{nogar02@cs.huji.ac.il}
\and
Michael Schapira \\ 
\huji  \\
\authemail{schapiram@cs.huji.ac.il }
\and
Aviv Tamar \\ 
\technion \\
\authemail{avivt@technion.ac.il}
}

\maketitle

\begin{abstract}
Recently, deep learning has been successfully applied to a variety of networking problems. A fundamental challenge is that when the operational environment for a learning-augmented system differs from its training environment, such systems often make badly informed decisions, leading to bad performance. We argue that safely deploying learning-driven systems requires being able to determine, in real-time, whether system behavior is coherent, for the purpose of defaulting to a reasonable heuristic when this is not so. We term this the \emph{\ProbName\ problem} (\ProbNameShort). We present three approaches to quantifying decision uncertainty that differ in terms of the signal used to infer uncertainty. We illustrate the usefulness of online safety assurance in the context of the proposed deep reinforcement learning (RL) approach to video streaming. While deep RL for video streaming bests other approaches when the operational and training environments match, it is dominated by simple heuristics when the two differ. Our preliminary findings suggest that transitioning to a default policy when decision uncertainty is detected is key to enjoying the performance benefits afforded by leveraging ML without compromising on safety.

\end{abstract}

\sloppy 
\section{Introduction}\label{sec:intro}

A recent trend in networking research is employing machine learning (ML) and, more specifically, deep learning (DL), to inform decision making. Example application domains include video streaming~\cite{oboe, mao2017neural, sun2016cs2p}, traffic management~\cite{chen2018auto,valadarsky2017learning}, resource scheduling~\cite{mao2019learning}, packet classification~\cite{liang2019neural},  caching~\cite{kirilin2020rl}, and congestion control~\cite{pantheon,jay2019deep,nie2019dynamic}. In all these contexts, the ability of deep neural networks (DNNs) to pick up on complex patterns in data has proved instrumental. 

Unleashing the power of DL on networking domains, however, requires care. DNNs often require many data samples to learn, and are so typically trained \emph{offline} and then put into practice. When the training environment is faithful to the operational environment, the DNN will likely perform well. However, DL is notorious for not \textit{generalizing} well when the two environments differ. One reason for such a mismatch is the standard practice of training on simulated/emulated environments~\cite{mao2017neural,jay2019deep}, which fail to capture the intricacies of real-world networks~\cite{puffer}. However, even if training occurs \textit{in situ} on real data, as advocated in~\cite{puffer}, the operational environment encountered after training might still greatly differ from the training environment due to variability in network conditions not adequately covered by the finite training data, or the introduction of new factors such as routing changes, network failures, the addition/removal of traffic sources, etc.

Can the benefits of ML-driven networking when training data is sufficiently rich be reaped without suffering the performance costs of bad generalization? We propose a general methodology for the safe deployment of DL-based systems: building into the system the means to detect, in real-time, scenarios it was not trained for and cannot make reliable decisions in. We term this challenge the \emph{\ProbName\ problem} (\ProbNameShort). Online safety assurance can accommodate the safe deployment of learning-augmented systems by facilitating switching to a reasonable default algorithm/heuristic (e.g., a widely deployed system that has already been ``tested in battle'' in many operational environments) when the learning-based system's decisions no longer seem sensible. In this sense, online safety assurance is intended to effectively serve as a ``safety net'' for DL-augmented systems.  %Importantly, since the system itself is data-driven, we cannot manually define the safety net, but must \emph{learn it from the data} as well.

%\ProbNameShort\ is related to the novelty detection (ND) problem~\cite{Scholkopf:2001:ESH:1119748.1119749,pimentel2014review,Tax:2004:SVD:960091.960109} in ML, which concerns detecting when observed data samples are ``too different'' from some input data set. In contrast to conventional ND, however, \ProbNameShort\ pertains to \textit{sequential} decision making, a topic that has received far less attention in the literature.
% and, to the best of our knowledge, is the first investigation of decision uncertainty in this setting. 

A key question when tackling \ProbNameShort, is how to best measure decision uncertainty. To ground this question in a formal framework, we consider the standard model for sequential decision making, namely, decision making under a Markov decision process (MDP)~\cite{bertsekas1995dynamic}. We observe that there are three natural signals within this model that could potentially be indicative of uncertainty: uncertainty regarding the environment state, incoherent choices of actions, and incoherent perceptions of how choices of actions impact performance. We present three notions of uncertainty, corresponding to these different signals, and present a methodology for how these can be utilized for online safety assurance. To the best of our knowledge, ours is the first study to explore uncertainty signals for safe sequential decision making.

To illustrate our approach, we ground our investigation in a concrete application---online bitrate adaptation (ABR) in video streaming---a concrete DL-based ABR algorithm---Pensieve~\cite{mao2017neural}---and a concrete default ABR algorithm---Buffer-Based (BB)~\cite{bb}. Our preliminary evaluation results corroborate the findings in~\cite{puffer}, showing that while Pensieve dominates BB when its training data and test data come from the same distribution, it is typically worse when the two differ, sometimes by a large margin. This motivates our approach by suggesting that incorporating into Pensieve the ability to switch to BB when decision uncertainty is detected can \textit{simultaneously} achieve high performance when Pensieve's training and operational environments match \textit{and} safety when the two differ. Our results for safety-assurance-enhanced variants of Pensieve provide evidence that this is indeed the case. Furthermore, our findings reveal that the choice of uncertainty signal can dramatically affect performance.

We believe that online safety assurance can facilitate the safe adoption of DL-augmented systems in many domains, and outline exciting directions for future research.

% At present, RL is too slow to learn online, and most scenarios train agents offline, obtain well-performing policies, and apply them in the online setting without further learning. 

\section{Online Safety Assurance}

We hypothesize that safe deployment of learning-based systems can be facilitated by detecting uncertainty in system decisions and defaulting to a safer policy when a system is outside its ``comfort zone''. The key question, however, is how to measure uncertainty in sequential decision making. 

To ground this question in a formal framework, we consider the standard model for sequential decision making---decision making under a Markov decision process (MDP)~\cite{bertsekas1995dynamic}. We next describe \ProbNameShort\ in the MDP setting and propose several approaches to online safety assurance, which differ in terms of the signal used for detecting uncertainty.

\subsection{Sequential Decision Making}\label{subsec:MDP}

We consider an \textit{agent} that interacts with an environment. At any discrete time step $t=0,1,\dots$, the agent takes \textit{action} $a_t$ from a set of possible actions $A$. Following the agent's action, the environment, which at time $t$ is at \textit{state} $s_t$, transitions to a new state $s_{t+1}$ with probability $P(s_{t+1}|s_t,a_t)$. Environment states belong to some set of possible states $S$. A \textit{reward function} $r$ maps each state-action pair $(s,a)$ to an associated reward $r(s,a)$. The agent's goal is to select actions that lead to high rewards. A common objective for the agent is optimizing the $\gamma$\textit{-discounted expected return} $\mathbb{E}^\pi \left[ \sum_{t=0}^\infty \gamma^t r(s_t,a_t) \right]$, where the expectation is taken with respect to the agent's decision making strategy $\pi$, henceforth termed the \emph{policy}. In general, $\pi$ is a mapping from the agents observation history $h_t=s_0,a_0,\dots,s_t$ to a \textit{probability distribution} over actions. In this work, we do not limit ourselves to a particular objective for the agent, nor to a specific class of policies $\pi$.

Many algorithms for finding optimal policies for MDPs employ some form of \textit{value function} estimation~\cite{bertsekas1995dynamic,sutton2018reinforcement}. The value function $V^\pi$ maps each state $s$ to a prediction of the future benefit to the agent from visiting this state. Formally, when considering the discounted objective, $V^\pi$ is the expected discounted return when the agent (whose policy is $\pi$), starts at that state $V^\pi(s) = \mathbb{E}^\pi \left[ \left. \sum_{t=0}^\infty \gamma^t r(s_t,a_t) \right| s_0=s\right]$.

The above formulation of decision making is very general and encompasses essentially all learning-based algorithms proposed for networking problems that we are aware of (including~\cite{mao2017neural,chen2018auto,mao2019learning,liang2019neural,oboe, jay2019deep, puffer}. In particular, this formulation encompasses algorithms that reflect reinforcement learning~\cite{sutton2018reinforcement} (RL), adaptive control~\cite{aastrom2013adaptive}, or approximate dynamic programming~\cite{bertsekas1995dynamic} approaches.

\subsection{The \ProbNameShort\ Problem}\label{subsec:OSAP}

During training, a learning-driven agent observes data from some distribution $\mu_{train}$, where by data we mean the history of observed states, actions, and rewards. During testing, however, the agent might observe data from a \textit{different} distribution $\mu_{test}$. The reason that the training and test distributions might differ can be a consequence of various reasons: the system being trained in simulation/emulation~\cite{puffer}, variability in real-world network conditions not adequately covered by the finite training data, or the learned decision making policy driving the agent at test time to environment states that were not explored during training. 

Since the learning-based components of the agent are trained on specific data, it is only natural that the farther away the agent is from its training regime, the less accurate the implicit or explicit information on which its decisions are based is.\footnote{We remark that the standard notion of \textit{generalization} in ML, and established generalization guarantees, are only applicable when the training and test distributions are the same. When outside the training distribution, as explored here, there are no generalization guarantees~\cite{ben2007analysis,pan2009survey,arjovsky2019invariant}.} How can we identify when the agent's decisions are not reliable due to a mismatch between its training and test distributions? In \ProbNameShort, given access to the agent's training data, the goal is to devise a rule that, based on the agent's test observation history $h_t$, decides whether the agent's next decision $\pi(h_t)$ is reliable or not.

\subsection{What to Measure?}\label{subsec:uncertainty-what}

To the best of our knowledge, a standard method for measuring decision uncertainty in sequential decision making has not yet been established. We take a step in this direction by proposing and comparing three approaches to quantifying decision uncertainty, which differ in the observed signal of uncertainty: uncertainty regarding the environment's states, the policy, and the value function. We refer to uncertainty estimation pertaining to these key MDP terms as U$_{S}$, U$_{\pi}$, and U$_{V}$, respectively.

\noindent{\bf U$_{S}$.} As the agent constantly observes the changes in the environment's state, a natural signal for uncertainty is a dissimilarity between the observed sequences of states in the training and test data. Deciding whether a test sample is too different from some training data (a.k.a., is an outlier, or out-of-distribution) is a standard (\emph{unsupervised}) ML task termed \textit{novelty detection} (ND)~\cite{Scholkopf:2001:ESH:1119748.1119749,Tax:2004:SVD:960091.960109}. U$_{S}$ is a natural extension of traditional ND to sequential decision making, and amounts to deciding whether the history of observed states in test is an outlier with respect to its observations on the training distribution. %In our considered model of decision making under MDP, this notion of ND corresponds to detecting a change in the state-transition probabilities $P(s_{t+1}|s_t,a_t)$ or policy $\pi(a_t|h_t)$ between training and test, as both can cause the agent to experience a different state distribution. 
% in the training distribution and the testing distribution are the same. 
Uncertainty in U$_{S}$ pertains to the agent's \emph{input}, that is, the observed environment state.

%Intuitively, a safety net based on a well-performing novelty detection method is conservative -- it always defaults when encountering unknown states. In this work, however, we also explore potentially less conservative approaches, as we describe next.

\noindent{\bf U$_{\pi}$ and U$_{V}$.} Unlike U$_{S}$, U$_{V}$ and U$_{\pi}$ measure uncertainty with respect to the agent's \emph{output}, i.e., its decisions. The rationale is that, ultimately, it is the output we wish to be certain of. In principle, even if the training and test distributions are different, the learned policy might still provide high performance in test. Thus, measuring uncertainty in the output could potentially prevent the agent from defaulting when this is not beneficial. Alternatively, U$_{V}$ and U$_{\pi}$ might potentially detect uncertainty in policy/value even when the training and test distributions are \textit{the same} (e.g., due to suboptimality of the learning algorithm), triggering transition to a safer, better performing policy.

\subsection{How to Measure?}\label{subsec:uncertainty-how}

\noindent{\bf Measuring U$_{S}$ as novelty detection.} We choose to use a classic ND method, namely, the one-class support vector machine (OC-SVM)~\cite{Scholkopf:2001:ESH:1119748.1119749}. OC-SVM enables learning a function that outputs $+1$ in a small region capturing most of the data points, and $-1$ elsewhere. See formal exposition in~\cite{Scholkopf:2001:ESH:1119748.1119749}.

Methods for measuring uncertainty in the output of learned functions have been developed for a variety of specific learning models~\cite{seber2012linear,rasmussen2003gaussian,auld2007bayesian,guo2017calibration}. We adopt a general approach that is applicable to arbitrary learning models, including deep neural networks, and can be applied to quantify both value and policy uncertainty. Our method is based on training an \textit{ensemble} of functions in the same training environment, and quantifying uncertainty in terms of the extent to which the outputs of the different functions are in agreement~\cite{osband2018randomized,sekar2020planning}.

% \vspace{0.1in}\noindent{\bf U$_{\pi}$ and U$_{V}$:} 
\noindent{\bf Measuring U$_{\pi}$ via agent ensembles.} Consider an ensemble of $i$ different agents trained in the same training environment, where the only difference in the training process is the initialization of the neural network variables. Intuitively, given a state $s_t$ and an observation history $h_t$, if each agent is certain in the output of its individual policy $\pi^i$, it is expected that the resulting probability vectors over actions outputted by different agents be similar. This would indicate that the policy learned by each agent is not highly dependent on how the neural network was initialized. To quantify similarity between probability vectors we use the Kullback–Leibler (KL) divergence~\cite{kullback1951information}, as in many previous studies~\cite{schulman2015trust,sekar2020planning}. Given probability distributions over actions $\bar{a}_1,\ldots,\bar{a}_i$ outputted by the $i$ agents comprising the agent-ensemble, the uncertainty according to U$_{\pi}$ is the sum of distances (in terms of KL divergence) of these outputs from the distribution over actions $\bar{a}$ obtained by taking the average over all agent outputs.

\noindent{\bf Measuring U$_{V}$ via value-function ensembles.} Many algorithms for finding optimal policies for MDPs employ some form of value function estimation~\cite{sutton2018reinforcement}. The value function $V^\pi$ in such algorithms is often represented as a neural network and is trained by observing the rewards derived from selecting actions according to the agent's policy $\pi$ (on the agent's training data). Importantly, even if an agent does not explicitly estimate state values, a value function for that agent can still be trained \textit{externally} by observing the history of states, actions, and rewards resulting from the agent-environment interaction while training.

To measure uncertainty with regards to the value function, we train $i$ value functions on the training distribution $\mu_{train}$. As with agent ensembles, the difference in training between different value functions is in how the neural network is initialized. The level of agreement between the outputs of the different value functions $V^{(\pi, i)}$ is used as an indication of certainty regarding the state value. Given $i$ state values outputted by the $i$ value functions comprising the value-function ensemble, the uncertainty according to U$_{V}$ is the total sum of distances between these $i$ values and the average value.

% This methodology is meant to capture the distance between the different outputs, 

% To the best of our knowledge, this is the first usage of an ensemble of critics. % removed because Ian Osband's work suggested value function ensembles.

\subsection{Setting Thresholds for Defaulting}\label{subsec:thresholding}

\ProbNameShort\ advises the agent about its uncertainty in its current decisions. To ultimately decide whether to default to another policy or not, we must threshold the uncertainty estimate somehow. To avoid premature transitions to the default policy because of sporadic or noisy data points, our thresholding techniques (illustrated in the ABR context in Section~\ref{sec:eval}), incorporate two ideas (1) considering  \textit{sequences} of data points; feeding into the OC-SVM samples in the form of the $k$ last observed environment states, and examining the variance in U$_\pi$ and U$_V$ across the last $k$ time steps, for some predetermined $k>0$, and (2) only defaulting if uncertainty is detected $l$ consecutive times, for some predetermined $l>0$. Setting the threshold for defaulting thus involves configuring $k$, $l$, and also, for U$_\pi$ and U$_V$ (which are continuous measures, unlike the binary U$_S$), the bar above which the action/value is considered uncertain.

{\bf Setting the defaulting threshold involves inherent tension between optimizing performance when the training and test environments are similar and controlling the possible damage when this is not so.} Even when the training and test distributions are identical, uncertainty might still occasionally arise due to differences in the specific samples drawn from this distribution in training and test. Hence, if the threshold is set to be ``too low'', the agent will default to another policy often even when its learned policy is most relevant. In contrast, if the threshold is ``too high'', the agent might stick with its learned policy even when the circumstances no longer justify this. {\bf How the threshold is set should thus reflect the system's designer/operator desired balance between performance and risk.} Determining the defaulting threshold with respect to a specific learning-based agent, default policy, environment, and uncertainty estimation metric, is an empirical question. We revisit this challenge in the concrete example of ABR in Section~\ref{sec:eval}.

A key question when evaluating safety assurance is how to fairly compare approaches that are based on different signals for uncertainty. We address this through \emph{calibration}: in our experiments, online safety assurance with respect to U$_S$, U$_\pi$, and U$_V$ is calibrated to attain \textit{the same performance} when $\mu_{training}=\mu_{test}$, and is evaluated based on its performance on the test distribution. This is inspired by traditional ND literature, in which a standard evaluation method is to set the threshold to achieve a prescribed true positive detection rate (say, $95\%$), and report the true negative detection rate. Our calibration approach can be regarded as an adaptation of this common practice to sequential decision making, in which performance is quantified by achieved rewards. %Importantly, our calibration aligns with a practical approach: first set the expected performance under nominal conditions, and then control the potential damage in extremes. 

\section{Case Study: Video Streaming}\label{sec:eval}

We consider the case-study of adaptive bitrate (ABR) selection in video streaming, a subject of extensive recent attention~\cite{mpc,mao2017neural,puffer}. In ABR video streaming, videos are encoded in different resolutions (bitrates) and segmented into chunks of (roughly) equal duration. ABR algorithms at video clients determine at which bitrate to download each video segment based on their local observations about network throughput. Variability in network throughput can be detrimental to QoE, potentially driving ABR algorithms to undershoot when selecting resolutions, overshoot and suffer video rebuffering, or change the bitrate too often. 

Recently, applying DL to ABR in various forms has been proposed (deep reinforcement learning in~\cite{mao2017neural} and deep network throughput prediction in~\cite{puffer}). In our ABR case study, we use Pensieve~\cite{mao2017neural} as the learned policy and Buffer-Based~\cite{bb} (BB) as the default (``safe'') policy.

We chose Pensieve for two reasons: (1) While Pensieve performs well when its training environment reflects the test environment~\cite{mao2017neural}, it suffers from bad performance when the two differ~\cite{puffer}, undermining its practical deployability and motivating the need for online safety assurance. (2) Pensieve employs an actor-critic~\cite{mnih2016asynchronous} reinforcement learning algorithm, and so has built-in reward optimization and value estimation (obviating the need for quantifying these externally to the protocol). BB is a simple ABR strategy that performs remarkably well in practice across a variety of network conditions~\cite{puffer} and is thus a suitable default policy.

\subsection{Evaluation Framework}\label{subsec:eval_framework}

\noindent{\bf Datasets.} For real-world data, we use two publicly available datasets: a 3G/HSDPA mobile dataset collected in Norway~\cite{norwaydataset} (we use the data generated in~\cite{mao2017neural}), and a 4G/LTE mobile dataset collected in Belgium~\cite{belgiumdataset}. For both datasets, $70$\% of the data was used for training, while the remaining $30$\% was used for testing. Validation was done on $30$\% of the training set.
% For the 4G/LTE dataset, \NR{add something about validation?}\MS{What about 4G/LTE? Text missing.} 
In addition, we generated $4$ synthetic datasets by sampling network throughput i.i.d. from different distributions: Gamma with shape $1$ and scale $2$, Gamma with shape $2$ and scale $2$, Logistic with $\mu=4$ and scale $0.5$, and Exponential with scale $1$.

\noindent{\bf Network emulation.} We build upon the experimental framework in~\cite{mao2017neural}, where MahiMahi~\cite{netravali2015mahimahi} is used to emulate network conditions from input network traces, with a $80$ms RTT between video client and server.

\noindent{\bf QoE metric.} We consider the conventional linear QoE metric from previous studies~\cite{mpc,mao2017neural}:
\[QoE = \sum_{n=1}^N R_n - \mu \sum_{n=1}^N T_n - \sum_{n=1}^{N-1} |R_{n+1} - R_n|,\]
where $N$ is the number of chunks in the video, $R_n$ is the bitrate at which video chunk $n$ was downloaded, and $T_n$ is the rebuffering time that resulted from downloading chunk $n$ at
bitrate $R_n$. Intuitively, optimizing this QoE metric corresponds to maximizing the average bitrate (first term) while minimizing rebuffering time (the second term) and the jitter between different bitrates (the third term).

\noindent{\bf Video.} We use the "EnvivioDash3" video from the DASH--246 JavaScript reference client~\cite{video} (the same video used in~\cite{mao2017neural}). The video is encoded in six different bitrates ($\{240, 360, 480, 720, 1080, 1400\}$p), and is divided into forty-eight video chunks, each around $4$-seconds long. To prolong the duration of the video we created a new video by concatenating the original video five times.

\noindent{\bf Learned and default ABR algorithms.} We use Pensieve~\cite{mao2017neural} and Buffer-Based (BB)~\cite{bb}, as implemented in~\cite{mao2017neural}, as the learned and default ABR policies, respectively.

\noindent{\bf Online safety assurance schemes.} We generate three safety-enhanced variants of Pensieve by incorporating into Pensieve online safety assurance with respect to each of our uncertainty metrics, U$_S$, U$_\pi$, and U$_{V}$.

U$_S$ is determined by a OC-SVM (see Section~\ref{subsec:uncertainty-how}) implemented using the SciPy package~\cite{2020SciPy-NMeth}. During test, at each time step $t$, the mean and standard deviation of the $10$ most recent network throughputs are calculated, and a sample consisting of the $k$ latest [mean, deviation] pairs is fed into the (trained) OC-SVM model. 
% a sample in the form of the $k$ latest [mean, deviation] pairs. 
In our experiments, $k=5$ for the empirical distributions and $k=30$ for the synthetic distributions (we observed that a longer window was required for good performance on the synthetic data, which we attribute to the high variance in these distributions). Given such a sample, the OC-SVM model provides a binary answer, classifying the sample as either in-distribution or out-of-distribution (OOD). When samples are classified as OOD for $l=3$ consecutive time steps, the system defaults to BB.

To compute U$_\pi$ and U$_V$, an ensemble (of agents and value functions, respectively) of size $i=5$ is trained. At each time step $t$, the two outputs (actions/values) whose distance from the average is highest are discarded and U$_\pi$ and U$_V$ are computed with respect to the three surviving outputs as explained in Section~\ref{subsec:uncertainty-how}. The system defaults to BB when the variance of this value across the last $k=5$ time steps exceeds a certain threshold $\alpha$ for $l$ consecutive times. $\alpha$ and $l$ for each of the two schemes are determined as explained below.

We henceforth refer to the safety enhancements that use U$_S$, U$_\pi$, and U$_V$, as \textit{novelty detection} (ND), \textit{agent ensemble} (A-ensemble), and \textit{value ensemble} (V-ensemble), respectively.

\noindent{\bf Threshold calibration.} To allow for fair comparison (see Section~\ref{subsec:thresholding}), we calibrate the defaulting thresholds for the evaluated U$_\pi$-based and U$_V$-based safety assurance schemes to match the performance (QoE) of the U$_S$-based scheme with respect to each considered training distribution. We defer the thorough investigation of how different thresholding strategies impact performance to future research.

\noindent{\bf Remark: offline and online running times.} Our safety assurance schemes require \textit{offline} training of novelty detection schemes (for U$_S$, OC-SVM in our case) or RL agents and value functions (for U$_\pi$ and U$_V$, respectively) and also \textit{online} computation to determine whether to transition to the default policy.
Training, in our experiments, was executed (offline) on 24-core Intel(R) Xeon(R) CPU E5-2630 v3 @ 2.40GHz w. Nvidia Tesla M60 GPU machines, and required less than eight seconds for OC-SVM (for U$_S$-based safety assurance), approximately eight hours for an RL agent (for U$_\pi$-based safety assurance), and approximately four hours for a value function (for U$_V$-based safety assurance).
The online decision making required far fewer resources, and was therefore executed on a single core Intel(R) Core(TM) i5-8500 CPU @ 3.00GHz w. GeForce GT 730 GPU machine. Each decision required approximately $0.5$ms for U$_S$-based safety assurance, $3$ms for U$_\pi$-based safety assurance, and $4$ms for U$_V$-based safety assurance. Since ABR decisions typically occur at the granularity of seconds, these running times are orders of magnitude lower than needed for accommodating timely decisions.

% \begin{figure}[h]
%     \centering
%     \includegraphics[width=0.48\textwidth]{figures/pensieve_vs_bb.png}
%     \caption{Pensieve \emph{vs.} BB when the training and test distributions are the same.}
%     \label{fig:pensieve_vs_bb}
% \end{figure}

% \subsection{Pensieve outperforms BB when the $\mu_{train}=\mu_{test}$}
\subsection{Pensieve with safety assurance still outperforms BB in-distribution}

\begin{figure}[h]
    \centering
    \includegraphics[width=0.48\textwidth]{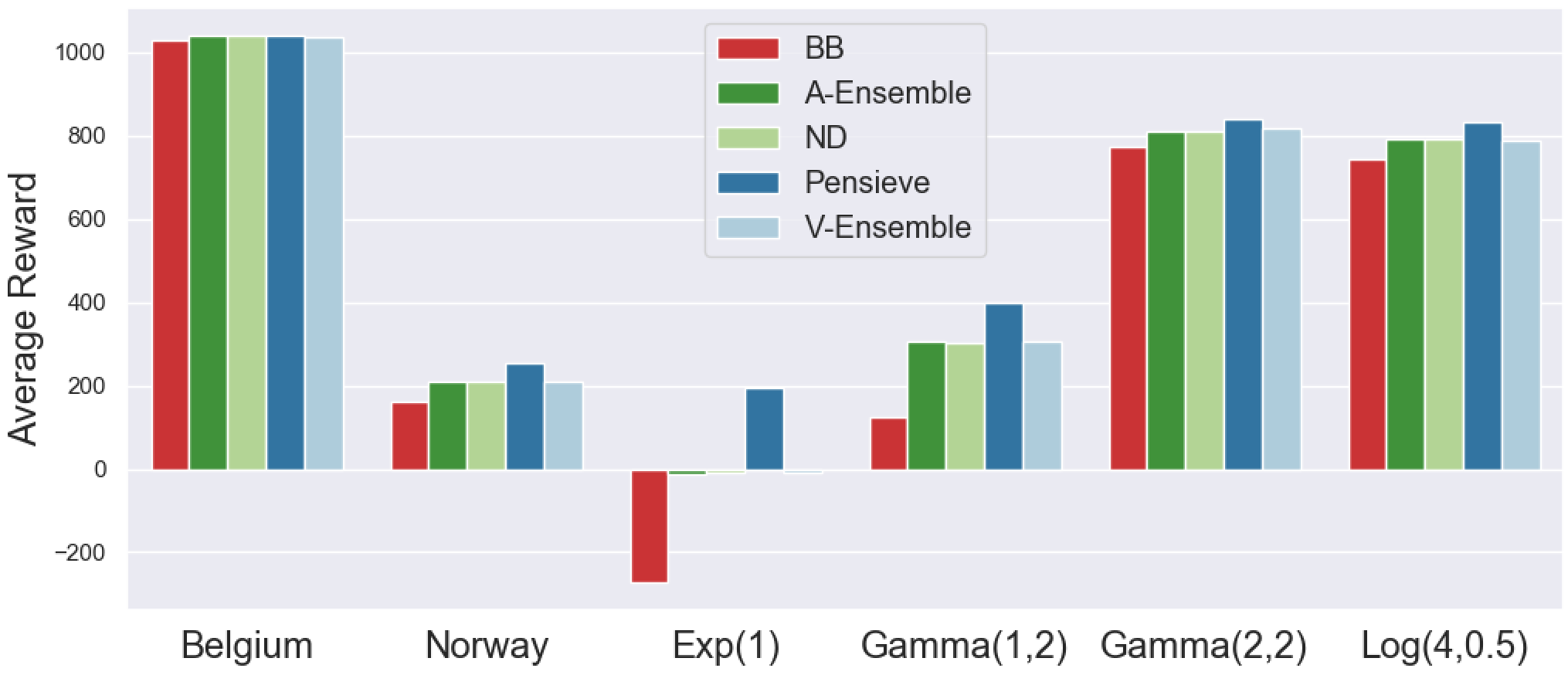}
    \caption{Pensieve with and without safety assurance \emph{vs.} BB when the training and test distributions are the same.}
    \label{fig:in-distribution}
\end{figure}

We first consider the performance of our safety-enhancements to Pensieve when \textit{in-distribution}, i.e., when the training and test datasets come from \textit{the same} distribution. \autoref{fig:in-distribution} plots the performance in test of ``vanilla'' Pensieve (with no safety assurance), ND, A-ensemble, V-ensemble, and BB, for all six (training,test) pairs of datasets that come from the same distribution. As expected, Pensieve, whose training environment here faithfully captures its test environment, outperforms BB in all experiments. Recall that the defaulting thresholds for the agent and value ensembles (denoted by A-ensemble and V-ensemble, respectively) are calibrated so as to match the performance of ND (with the fixed thresholding strategy described in Section~\ref{subsec:eval_framework}) on the training distribution. Therefore, when in-distribution, the three schemes always achieve the same level of performance, as seen in the figure. Observe that this level of performance is higher than that provided by BB and lower than that provided by Pensieve. As discussed in Section~\ref{subsec:thresholding}, the latter is a necessary price for accomplishing safety when \textit{out-of-distribution} (OOD). Different choices of thresholds would strike different trade-offs between performance in distribution and safety when out of distribution (see Section~\ref{subsec:thresholding}).

% Figure of Pensieve vs. Buffer-Based (only) on all 6 distributions (train=test). Not normalized by random.

% \subsection{Pensieve is dominated by BB when $\mu_{train}\neq \mu_{test}$}\label{subsec:test_not_train_res}
\subsection{Pensieve is dominated by BB when out-of-distribution}\label{subsec:test_not_train_res}
% Figure of Pensieve vs. Buffer-Based vs. Random (only) on the remaining 42 pairs (max/min/mean/median)

\begin{figure}[h]
    \centering
    \begin{subfigure}{0.48\textwidth}
        \includegraphics[width=\textwidth]{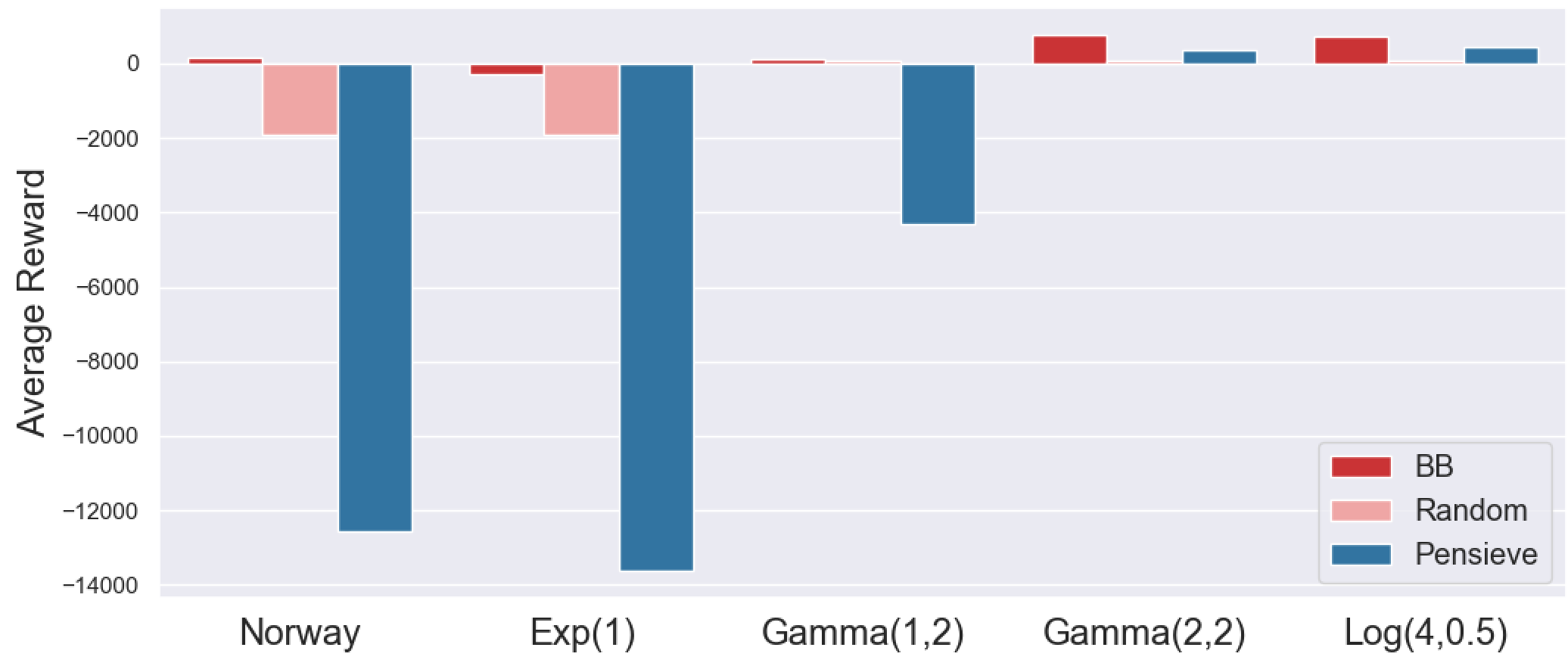} \label{subfig:belgium}
        \caption{Pensieve trained on Belgium and evaluated on the other datasets.}
    \end{subfigure}
    \begin{subfigure}{0.48\textwidth}
        \includegraphics[width=\textwidth]{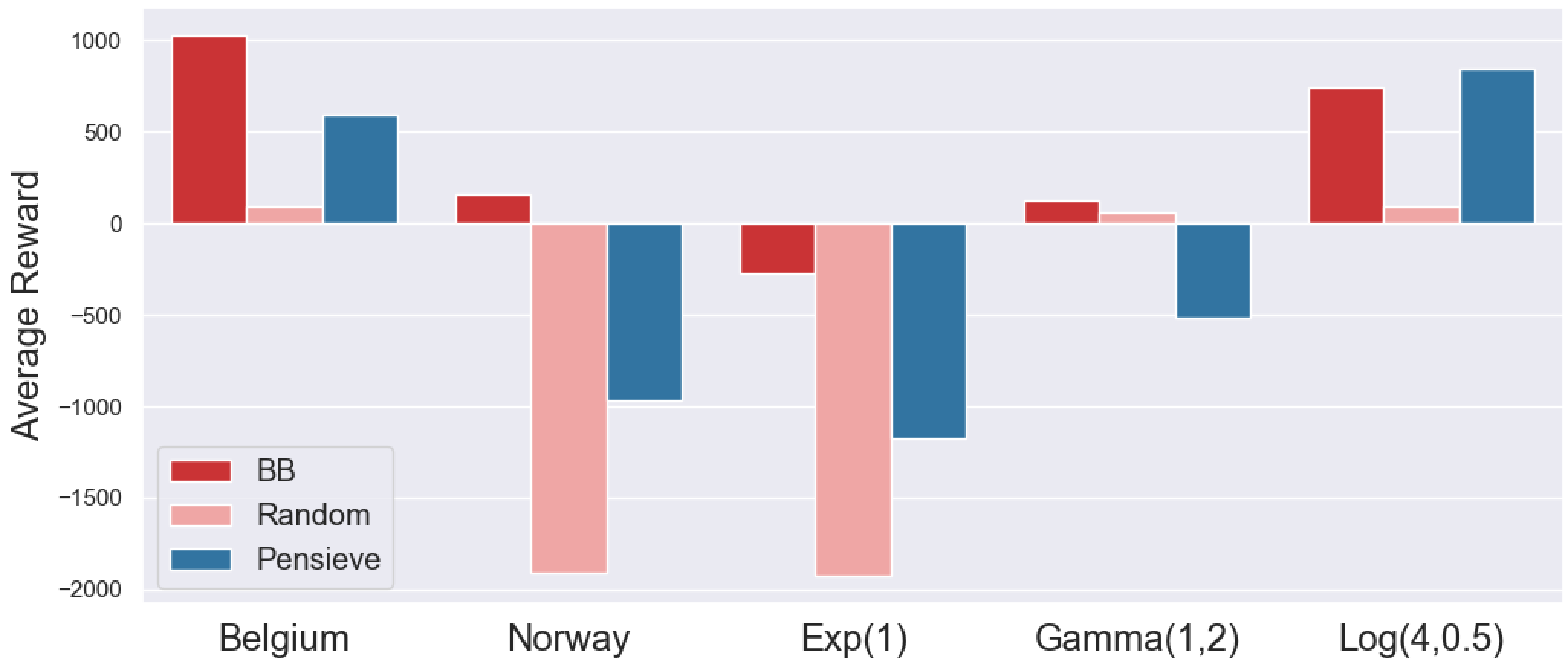}\label{subfig:gamma22}
        \caption{Pensieve trained on Gamma(2,2) and evaluated on the other datasets.}
    \end{subfigure}
    \caption{Illustration of Pensieve's (problematic) generalization to other environments}
    \label{fig:contrast_random}
\end{figure}

We next show that when evaluated on a test distribution that does not match its training distribution, Pensieve can fail catastrophically, in some cases even performing worse than a naive baseline that always selects the next bitrate \textit{uniformly at random}, termed ``Random'' henceforth. To illustrate this, \autoref{fig:contrast_random} contrasts the performance (QoE) achieved by Pensieve, BB, and Random for two training distributions: Belgium and Gamma(2,2). Observe that, with but one exception, Pensieve is outperformed by BB, and is sometimes even (significantly) outperformed by Random. This attests to Pensieve's inability to generalize well, as also shown empirically in~\cite{puffer}.

\begin{figure}[h]
    \centering
    \includegraphics[width=0.475\textwidth]{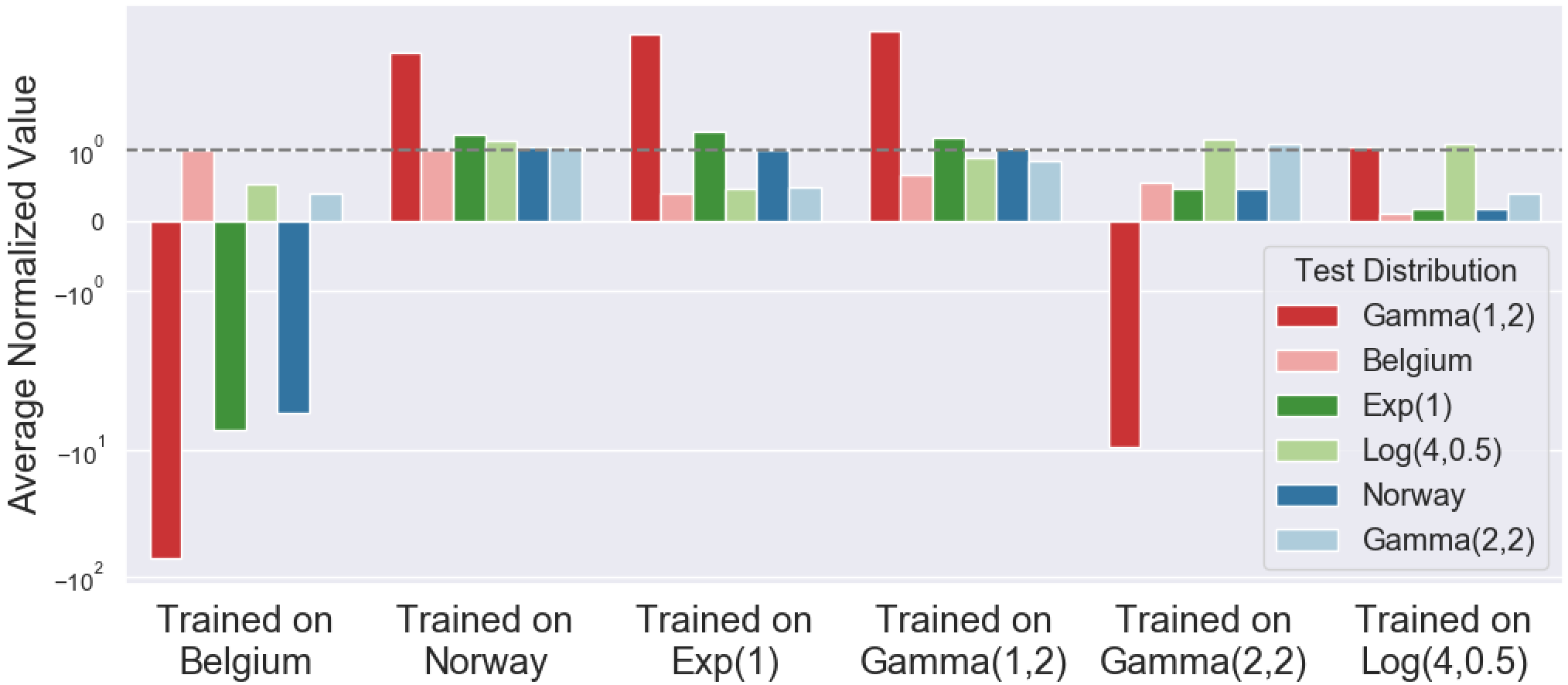}
    \caption{Pensieve's performance across all datasets.}
    \label{fig:pensieve_all}
\end{figure}

Hereafter, to clearly compare Pensieve's (with and without safety assurance) and BB's performance across different distributions, and contend with QoE scores being sometimes positive and sometimes negative, we normalize performance as follows. In all subsequent figures, a performance value of $0$ corresponds to Random's performance (on the relevant dataset), whereas a performance of $1$ corresponds to the gap between BB's performance and Random's performance. \autoref{fig:pensieve_all} depicts Pensieve's normalized scores when trained on different distributions and evaluated on \emph{other} distributions. When Pensieve's score is lower than $1$, this indicates that it is outperformed by BB, whereas when its score is lower than $0$, it is outperformed also by Random. Note that the scale on the $y$-axis is linear in the region [-1,1] and \textit{log scale} elsewhere. As can be seen in \autoref{fig:pensieve_all}, Pensieve is typically outperformed by BB when OOD.

\begin{figure}[h]
    \centering
    \includegraphics[width=0.475\textwidth]{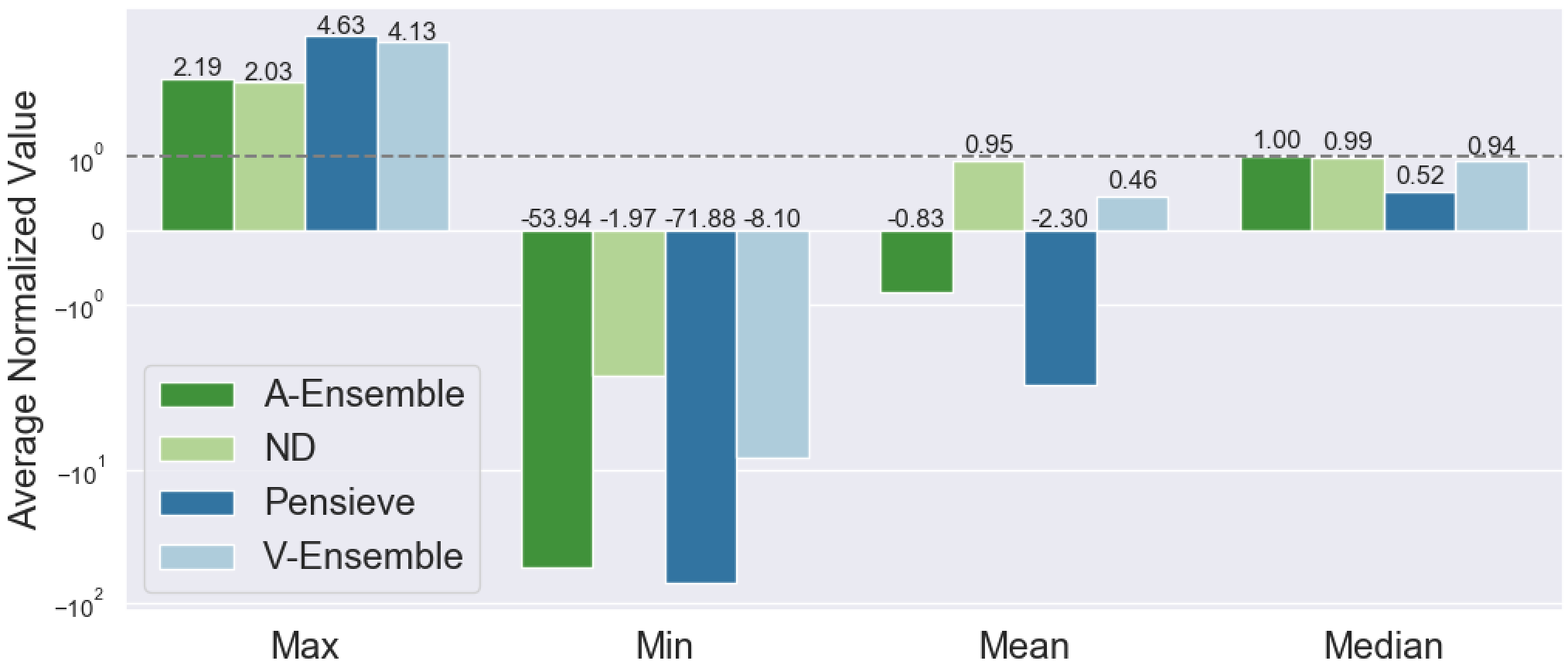}
    \caption{Comparison of different safety-enhanced variants of Pensieve OOD.}
    \label{fig:bb_dominates_pensieve}
\end{figure}

\subsection{Contrasting the three safety assurance schemes when OOD}
% \subsection{Contrasting three online safety assurance schemes}

\begin{figure}[h]
    \centering
    \includegraphics[width=0.48\textwidth]{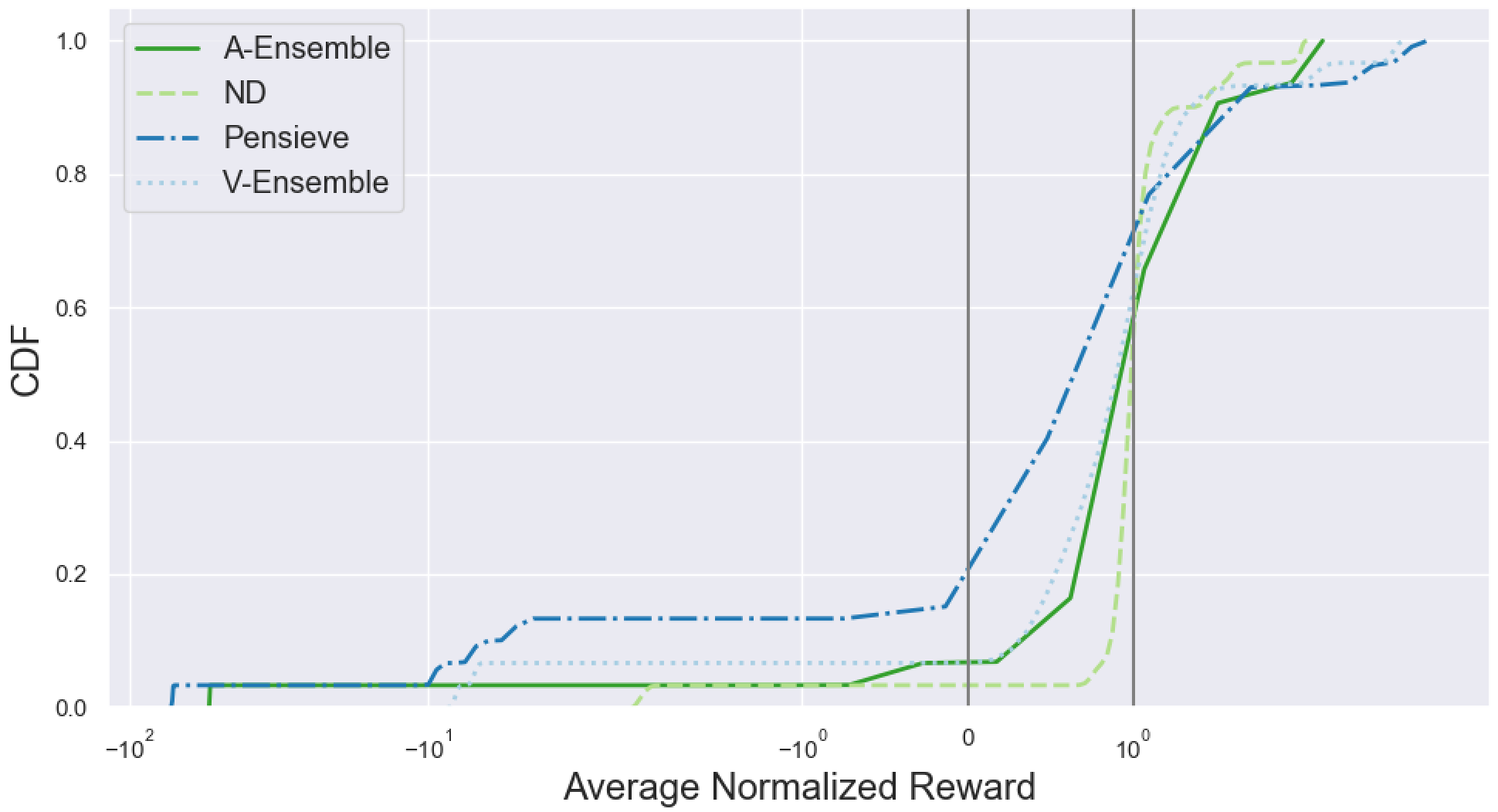}
    \caption{A CDF of performance for the different schemes across all $30$ training-test combinations where test is OOD}
    \label{fig:cdf}
\end{figure}

\autoref{fig:bb_dominates_pensieve} depicts the normalized maximal, minimal, mean, and median performance of the three safety-enhanced Pensieve variants across all $30$ combinations of training and test datasets that \textit{do not} come from the same distribution. We also present the performance of Pensieve (without safety assurance). Again, note the log scale outside [-1,1]. To provide more visibility into our results, \autoref{fig:cdf} presents a CDF of performance across our experiments for all $30$ training-test dataset pairs.

Evidently, all three safety assurance schemes are better than vanilla Pensieve in terms of min, mean, and median values. \autoref{fig:bb_dominates_pensieve} shows that the choice between the different safety assurance schemes allows flexibility in trading off risk (min performance) for payoff (max/mean/median performance). How to balance the different factors is a choice to be made by the system designer. In the following, however, we seek general claims about the risk/payoff trade-off for each scheme.

Interestingly, {\bf A-ensemble is essentially dominated by the other two safety assurance approaches}, with significantly lower min value, worse than Random (!) mean value, marginally better median value than the other two, and comparable (to ND) or worse (than V-ensemble) max performance. 
We conjecture that this is due to the fact that when training an ensemble of agents, though the training domain is the same for all, each might learn a \textit{different} (good) policy. Consequently, variability between agent outputs might manifest \textit{even on the training distribution}. Since we calibrate performance on training, this variability effectively leads to setting the threshold for defaulting for A-ensemble to be more tolerant to policy disagreement, and consequently failing to detect when an agent is operating OOD. This suggests that \textbf{due to the high variability in training, A-ensembles are inherently less reliable uncertainty estimators for \ProbNameShort.} 

Deciding between ND and V-ensemble depends on the specific criterion; V-ensemble is better in terms of max performance, whereas ND is better in terms of the min and mean performance (the two are comparable in terms of median). Thus, {\bf ND constitutes a safer choice, whereas V-ensemble can potentially provide higher performance gains.} 
To explain this, first note that {\bf V-ensembles do not suffer from the variability of A-ensembles as they are trained with respect to \textit{a single} agent's policy}. Second, {\bf ND reflects the most conservative approach, as it defaults whenever OOD samples are detected, while V-ensemble may potentially yield higher rewards when choosing to stick with Pensieve on OOD samples when the value has high certainty.} 

\section{Related Work}\label{s:related_work}

The mismatch between training and test distributions is the main challenge in \emph{off-policy RL}~\cite{levine2020offline}. Theoretical sample complexity guarantees were explored in~\cite{thomas2015high,thomas2015highim,ghavamzadeh2016safe,thomas2016data,hanna2017bootstrapping}, and recent work explores offline deep RL~\cite{sarafian2018safe,fujimoto2018off,kumar2019stabilizing}. In these works, the mismatch is due to a change in policy, \textit{not} in state transitions, as in our context. Perhaps surprisingly, generalization in RL has received relatively little attention~\cite{taylor2009transfer}, and recent studies focus on procedurally generated domains such as mazes~\cite{tamar2016value,zhang2018study,juliani2019obstacle} or platform games~\cite{nichol2018gotta}, and learning robust policies~\cite{huang2017adversarial,pattanaik2018robust}. None of these studies addresses the question of how to safely react to a change in the environment.

Classical methods for novelty detection include OC-SVM~\cite{Scholkopf:2001:ESH:1119748.1119749} and support vector data description~\cite{Tax:2004:SVD:960091.960109}. Recent methods build on deep generative models~\cite{an2015variational,schlegl2017unsupervised,wang2017safer,Song2018LearningNR,DBLP:journals/corr/abs-1901-03407}.  \emph{Curiosity-based exploration} in RL involves guiding the agent to unvisited states \emph{during training}~\cite{oudeyer2007intrinsic,schmidhuber2010formal}. Studies of curiosity-based exploration measure uncertainty in models of the environment~\cite{houthooft2016vime,pathak2017curiosity}, in state observations~\cite{bellemare2016unifying,burda2018exploration}, and in the level of agreement between an ensemble of value functions or policies~\cite{osband2016deep,osband2018randomized,sekar2020planning}. Our uncertainty estimation methods are inspired by this line of research, but strive to achieve a very different goal: \emph{avoiding uncertainty at test time}. Consequently, our challenges and solutions are different.
\section{Conclusion and Directions for Future Research}\label{sec:conclusion}

Learning-augmented systems have received much attention of late. However, releasing deep-learning-based systems in the wild requires guaranteeing that these attain acceptable performance even when their operational environment deviates from their training environment. We proposed detecting, in real time, when the decisions reached by a system are no longer unreliable, and defaulting to a safer alternative. Our initial investigation of such online safety assurance for ABR video streaming revealed that applying novelty detection methods or value-based uncertainty estimation for this purpose is promising, whereas using action-based uncertainty as a signal fares much worse. Natural directions for future research include the further investigation of which of the two online assurance schemes, novelty-detection-based and value-uncertainty-based, is more appropriate in different contexts, and exploring the implications for the performance of different thresholding strategies. Other intriguing research directions include extending our preliminary findings for ABR by considering other DL-based ABR systems (e.g.,~\cite{puffer}) and default policies, and investigating online safety assurance when training is performed in situ~\cite{puffer}. In addition, the exploration of online safety assurance in other application domains opens exciting directions for future research.

\section*{Acknowledgments}
Michael Schapira is partly funded by the Israel Science Foundation (ISF) and an NSF-BSF grant (BSF-2019798). Aviv Tamar is partly funded by the Israel Science Foundation (ISF-759/19) and the Open Philanthropy Project Fund, an advised fund of Silicon Valley Community Foundation.

\bibliographystyle{abbrv} 
\begin{small}
\bibliography{references.bib}
\end{small}
\end{document}